\documentclass[conference]{IEEEtran}
\IEEEoverridecommandlockouts
\usepackage{cite} 
\usepackage{amsmath,amssymb,amsfonts}
\usepackage{algorithmicx}
\usepackage{textcomp}
\usepackage{xcolor}
\usepackage{graphicx} 
\usepackage{algorithm} 
\usepackage{algpseudocode} 
\usepackage{tabularx}   
\usepackage{booktabs}   
\usepackage{array} 
\usepackage{svg}
\usepackage{graphicx}
\usepackage{subcaption}
\usepackage{caption}
\usepackage[capitalise]{cleveref} 
\def\BibTeX{{\rm B\kern-.05em{\sc i\kern-.025em b}\kern-.08em
    T\kern-.1667em\lower.7ex\hbox{E}\kern-.125emX}}

\begin{document}



\title{An Evolutionary Game-Theoretic Merging Decision-Making Considering Social Acceptance for Autonomous Driving}

\author{\IEEEauthorblockN{1\textsuperscript{st} Haolin Liu}
\IEEEauthorblockA{\textit{School of Mechanical Engineering} \\
\textit{Beijing Institute of Technology}\\
Beijing, China \\
haolin.liu@bit.edu.cn}
~\\

\and
\IEEEauthorblockN{2\textsuperscript{nd} Zijun Guo}
\IEEEauthorblockA{\textit{School of Mechanical Engineering} \\
\textit{Beijing Institute of Technology}\\
Beijing, China \\
zijun.guo@bit.edu.cn}

\and
\IEEEauthorblockN{3\textsuperscript{rd} Yanbo Chen}
\IEEEauthorblockA{\textit{School of Mechanical Engineering} \\
\textit{Beijing Institute of Technology}\\
Beijing, China \\
yanbo.chen@bit.edu.cn}

\and
\IEEEauthorblockN{4\textsuperscript{th} Jiaqi chen}
\IEEEauthorblockA{\textit{School of Mechanical Engineering} \\
\textit{Beijing Institute of Technology}\\
Beijing, China \\
3120230290@bit.edu.cn}

\and
\IEEEauthorblockN{5\textsuperscript{th} Huilong Yu*}
\IEEEauthorblockA{\textit{School of Mechanical Engineering} \\
\textit{Beijing Institute of Technology}\\
Beijing, China \\
huilong.yu@bit.edu.cn}
*Co-Corresponding author
~\\
\and
\IEEEauthorblockN{6\textsuperscript{th} Junqiang Xi*}
\IEEEauthorblockA{\textit{School of Mechanical Engineering} \\
\textit{Beijing Institute of Technology}\\
Beijing, China \\
xijunqiang@bit.edu.cn}
*Co-Corresponding author
~\\
}

\maketitle


\begin{abstract}
Highway on-ramp merging is of great challenge for autonomous vehicles (AVs),  since they have to proactively interact with surrounding vehicles to enter the main road safely within limited time. However, existing decision-making algorithms fail to adequately address dynamic complexities and social acceptance of AVs, leading to suboptimal or unsafe merging decisions. To address this, we propose an evolutionary game-theoretic (EGT) merging decision-making framework, grounded in the bounded rationality of human drivers, which dynamically balances the benefits of both AVs and main-road vehicles (MVs).
We formulate the cut-in decision-making process as an EGT problem with a multi-objective payoff function that reflects human-like driving preferences. By solving the replicator dynamic equation for the evolutionarily stable strategy (ESS), the optimal cut-in timing is derived, balancing efficiency, comfort, and safety for both AVs and MVs. A real-time driving style estimation algorithm is proposed to adjust the game payoff function online by observing the immediate reactions of MVs. Empirical results demonstrate that we improve the efficiency, comfort and safety of both AVs and MVs compared with existing game-theoretic and traditional
planning approaches across multi-object metrics.
\end{abstract}


\section{Introduction}
The growing deployment of AVs has improved traffic safety and efficiency through advanced decision-making systems. However, full commercialization remains limited by low social acceptance\cite{tang2021atac}. The core elements include trust in autonomous technology, perceived usefulness, and perceived safety, which are often measured by the extent to which human drivers are influenced by AVs, particularly in lane-merging scenarios\cite{yuan2016suppose, chen2025intention}. To enhance social acceptance, we need more accurate and interpretable modeling of interactive behaviors to generate more human-like driving behaviors.

\begin{figure}[t]
  \centering
  \includegraphics[width=\linewidth]{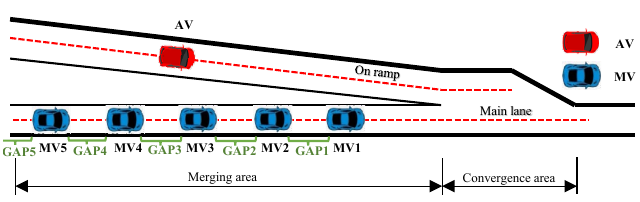}
  \caption{The lane-merging scenario.}
  \label{scenario}
\end{figure}

The lane-merging scenario is as shown in Fig.~\ref{scenario}, comprising a merging area and a convergence area. The AV on ramp needs to determine the appropriate merging gap before entering the convergence area. 
In this process, the yielding intention of each MV is unknown to the AV, which leads to misunderstandings such as insufficient safety guarantees and overly conservative decision\cite{wang2023social}. Proactive testing strategies are considered to interact with MVs and estimate their driving intentions, which presents two primary challenges: 
1) accurately modeling the dynamic interaction process and 2) balancing the benefits of AVs and MVs under unknown MV intentions.

To tackle the first challenge, prior approaches have predominantly relied on learning-based methods, optimization-based and game-theoretic methods. 
Irshayyid et al. proposed a DRL-based cooperative merging strategy using maskable PPO, optimizing energy, comfort, and traffic flow for platoon merging under lane reduction\cite{irshayyid2023comparative}.
Chen et al. developed FW-DBNs, a dynamic Bayesian network-based model that enhances real-time velocity prediction through adaptive window search and feedback weighting\cite{chen2024fw}. 
However, learning-based methods often suffer from distribution shifts and a lack of interpretability, which pose significant concerns for safety-critical applications. Chen et al. proposed a Branch MPC framework that handles multimodal reactive behaviors of uncontrolled agents using scenario and trajectory trees, and balances robustness and performance via risk-aware objectives\cite{chen2022interactive}.
Optimization-based methods, on the other hand, can easily fall into local optima due to the nonlinear system dynamics and non-convex constraints. 
In light of these limitations, game-theoretic frameworks have emerged as a promising alternative, offering a principled way to model strategic interactions among multiple agents in highly interactive scenarios \cite{reviewofgametheory}. 

In game theory, each game participant is characterized by a personalized utility function that incorporates both its own actions and those of other agents. The primary objective of each player is to strategically optimize its payoff function through iterative decision-making processes. 
Previous research has extensively explored equilibrium solutions for merging decision-making. Some approaches employ semantic-level decisions as game strategies, constructing a matrix game to identify Pareto-optimal Nash equilibria for discrete merging probabilities \cite{zhang2024automated, 9197129, 8765399}. While the classical Nash game fails to describe the dynamic adjustment process in which participants iteratively make optimal responses based on behaviors of opponents until reaching the Nash equilibrium. 
Other studies formulated the interaction between AV and MVs as a Stackelberg game with a leader–follower game structure \cite{wei2022game, 9261984}. In contrast to Nash equilibria that treats all game participant equally, defining leaders and followers in the real world is ambiguous, limiting modeling rationality and interpretability. 
Besides, both of them rely on the assumption of perfect rationality, which is difficult to satisfy in real world. 
To address these, evolutionary game-theoretic (EGT) methods have been proposed, which assumes game participants to be boundedly rational and use dynamic replicator equations to describe repeated interactions and dynamic strategy evolution \cite{ahmad2023applications, AsadiRecommendation}. These characteristics provide a viable framework to accurately model the interaction process.

Additionally, prior research suggests a potential correlation between driving style and intention, where aggressive drivers tend to prioritize efficiency and are more likely to resist cut-in maneuvers while conservative drivers often emphasize safety and comfort\cite{ren2019new}.
Most decision-making methods assume perfect Vehicle-to-Vehicle (V2V) communication to access driving intentions and driving styles of MVs, which is an impractical assumption in the real world\cite{molloy2017inverse}. To address this, online estimation of driving styles is necessary through historical motion states. 
However, existing methods based on inverse differential games \cite{bouzidi2025interaction} and inverse reinforcement learning\cite{wang2021socially} rely on the historical trajectory of target vehicles, lacking real-time dynamic interaction between the ego vehicle and target vehicles during estimation, which results in insufficient interaction sensitivity and risk adaptability \cite{jiang2025stackelberg}. Since EGT achieves dynamically stable equilibrium through repeated interaction, online estimation can rely on observing real-time interactive behaviors of human drivers rather than requiring advanced V2V communication systems.

In light of the challenges associated with modeling the interaction and balancing the benefits under unknown intentions, this work presents solutions tailored to tackle these issues. The original contributions are as follows: 
\begin{enumerate}
	\item[1)] We propose a dynamic game framework for decision-making in AVs. This framework designs the strategic evolution of game participants and solves for the ESS as the equilibrium strategy, while considering the unknown driving style of MVs in incomplete environments.
        \item[2)] We design an EGT algorithm that incorporates the bounded rationality of human drivers to formulate the interactive decision-making process between AV and MVs. By integrating a multi-objective payoff function, the algorithm derives the ESS to achieve adaptive and human-like cut-in behavior.
	\item[3)] An online algorithm is developed to estimate driving styles and adjust the multi-objective payoff function in real time. It dynamically observes and assesses the immediate reactions of MVs during interactions with AVs through continuous and interactive feedback on driving behavior.
\end{enumerate}

The rest of this paper is organized as follows. Section II describes the lane-merging scenario in mixed traffic and the dynamic game framework. Section III introduces the EGT merging decision-making algorithm. Section IV provides test results from data indicators. Finally, section V concludes this article and discusses future perspectives.
\section{Problem Formulation}
\begin{figure*}[!t]  
  \centering
  \includegraphics[width=\textwidth]{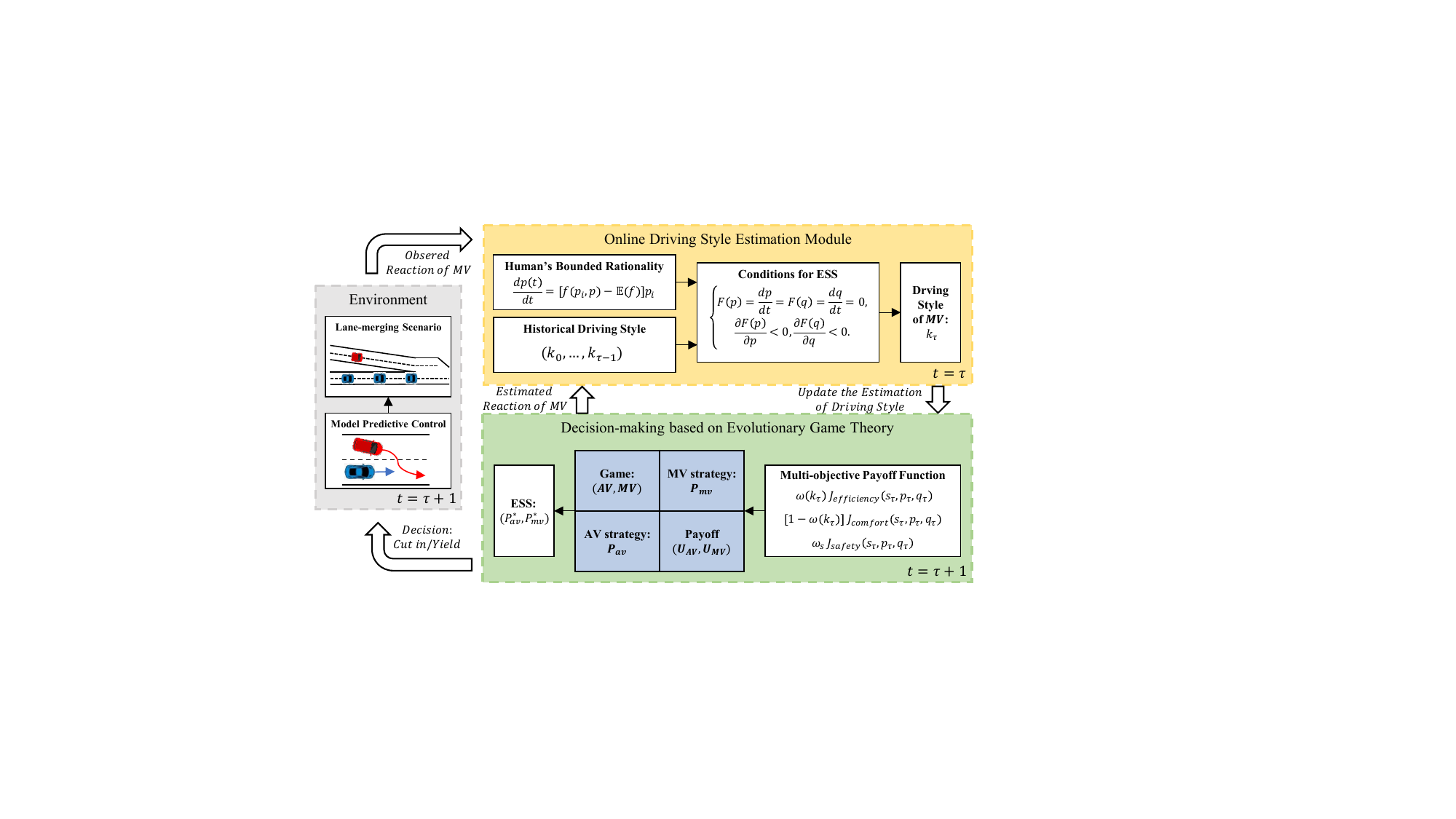} 
  \caption{The evolutionary game-theoretic decision-making framework.} 
  \label{framework}
\end{figure*}
\subsection{Lane-Merging Scenario}
This work focuses on a single-lane merging scenario, in which the AV sequentially interacts only with the adjacent MVs based on their priority in the queue, assuming that MVs do not perform lateral lane changes. This simplified modeling strategy is widely adopted in existing studies \cite{zhang2024automated, 9197129, 8765399} to control problem complexity and emphasize the features of the proposed decision-making method. We acknowledge that these idealized assumptions have limitations in real-world multi-lane, complex environments, and future work will extend the framework to address more general scenarios
In the scenario shown in Fig.\ref{scenario}, where five gaps are available, the AV is required to interact with surrounding MVs and choose an appropriate merging gap before entering the convergence area.

To this end, its decision-making system must comprehensively consider multiple factors, including the diverse behaviors of surrounding vehicles, the spatio-temporal characteristics of merging locations, as well as traffic efficiency, safety, and comfort during lane merging. In this process, AV interacts with MVs individually and sequentially according to their entry sequence. This interaction involves monitoring the kinematic states of each vehicle and interpreting turn signals, ultimately determining the appropriate merging gap and priority sequence to complete lane-changing maneuver at the convergence area. 

\subsection{Dynamic EGT framework}

The framework is described in Fig.\ref{framework}. The proposed framework consists of two tightly-coupled modules: (i) a EGT decision-making module, and (ii) an online driving style estimation module. 

Specifically, the proposed approach operates as follows.
Based on the bounded rationality of human drivers and their historical driving styles, we first solve the conditions to obtain the latest driving style estimation at time $ t=\tau$, thereby updating the multi-objective payoff function within the decision-making module. At the subsequent moment $ t=\tau+1$, a new game matrix is established based on this updated payoff function, and the ESS is calculated as the optimal strategy. Here, $ P_{av}^*$ serves as the selected strategy of AV, which is input into the lower-level MPC controller to guide the proactive actions of AV within the environment. Simultaneously, the real-time reaction of the MV to these proactive actions are observed. Additionally, at time $ t=\tau+1$, we estimate the real-time reaction of MV according to $ P_{mv}^*$ and compare it with the observed real-time reaction. This comparison is then used as input for the driving style estimation in the next step. By this iterative process, we achieve online updating of the driving style through proactive interactions until convergence is reached, enabling appropriate decision-making in an environment characterized by incomplete information.

\begin{algorithm}[t]
\caption{Overall Decision-making Framework}
\begin{algorithmic}[1]
\Require $\{x_i\}_{i \in N_0 \cup N_1}$
\Ensure $\{x_0^t\}_{t \in \{ 1, \ldots, T \}}$
\State Initialize $\{\xi_i\}_{i \in N_0 \cup N_1} $ and $\{k_i\}_{i \in N_0 \cup N_1}$
\While{$\xi_0 \leq \text{sizeof}(N_1)$}
    
    \If{$\exists \xi_j = \xi_0 + 1$}
        \State $k_j \gets \text{Driving\_style}(u_j^h)$
        \State $\{x_0^t\}_{t \in \{ 1, \ldots, T \}} \gets \text{EGT}(x_0, x_j, k_j)$
    \EndIf
    \State Wait for next planning phase
    \State Update $\{x_i\}_{i \in N_0 \cup N_1}$
    \State Update $\{\xi_i\}_{i \in N_0 \cup N_1} \gets \text{merging\_list}(x_i)$ and $\xi_0$
\EndWhile
\end{algorithmic}
\label{algorithm1}
\end{algorithm}

In Algorithm~\ref{algorithm1}, $ N_0 $ and $ N_1 $ are the sets of AVs and MVs. $ x_i$ represents the state of vehicle $ i$, and the final output is the planned trajectory $ x_0^t$ of AV within a decision-making horizon $ T$. Firstly, we initialize a priority queue $ \{\xi_i\}_{i \in N_0 \cup N_1} $ based on the temporal sequence of vehicles entering the lane-merging scenario and predefine driving style coefficients $ \{k_i\}_{i \in N_0 \cup N_1}$. $ \xi_i$ denotes the priority order of vehicle $ i$ in the set and $ \xi_0 $ indicates the priority order of AV. When the AV does not hold the highest priority in the queue, it performs an estimate of the driving style of the preceding vehicle $ j$ in the queue, engages in an evolutionary game, makes decisions, and generate planning trajectory for the next time step $ t $. The priority queue and vehicle states are iteratively updated until AV becomes the vehicle with the highest priority. It is worth noting that the driving style $ k_j $ requires analyzing vehicle's driving behavior over a historical period $ h$ (line 4 of Algorithm 1), and thus the algorithm must be executed sequentially in chronological order.

\section{Game-Theoretic Decision-Making}
In this section, we introduce details of the game players, strategies, online driving style estimation, payoffs and solver of the evolutionary game.
\subsection{Game Players and Merging Strategy}
In principle, we need to consider the interactions between AV and all MVs, each of which may exhibit lateral and longitudinal movements. However, this would lead to an exponential increase in the dimensionality of game matrix. Given that the computational complexity scales as $ O(n^2)$, computational overhead becomes prohibitive. To simplify the game matrix, this study introduces the following assumption.

\textit{Assumption 1}: After forming a priority queue based on the timing of entry into the merging area, the AV interacts only with adjacent vehicles in the queue.

Typically, researchers quantify interaction intensity using metrics such as the volume-to-capacity ratio of a roadway segment. In this merging scenario, we hypothesize that AV interacts solely with adjacent vehicles in the queue. This allows us to reduce the high-dimensional game matrix to a two-dimensional matrix involving only two participants.

\textit{Assumption 2}: During interactions, MVs maintain their lanes and only perform longitudinal movements, avoiding lateral maneuvers.

In real-world driving scenarios, when MVs encounter merging vehicles, drivers typically respond by adjusting speed rather than executing lateral lane changes. This behavior stems from the need to minimize collision risks and lateral movements could disrupt traffic in neighboring lanes and escalate accident probabilities. Accordingly, we construct the game matrix using semantic-level decision sequences. The decision set of AV can be defined as:
\begin{equation}
P_\mathrm{AV} = \left\{ Yield, Merge \right\}.
\label{decision set of AV}
\end{equation}

The decision set of MV can be defined as:
\begin{equation}
P_\mathrm{MV} = \left\{ Yield, Accelerate\right\}.
\label{decision set of MV}
\end{equation}

Based on these, the game matrix can be established as shown in the Table \ref{tab1}, 
\begin{table}[t]
\caption{MANTIC-LEVEL GAME MATRIX}
\label{tab1}
\centering
\renewcommand{\arraystretch}{1.5} 
\begin{tabularx}{0.45\textwidth}{|>{\bfseries}c|>{\centering\arraybackslash}X|>{\centering\arraybackslash}X|} 
\hline
\multicolumn{1}{|c|}{\textbf{AV/MV}} & 
\multicolumn{1}{c|}{\textbf{Yield(q)}} & 
\multicolumn{1}{c|}{\textbf{Accelerate(1-q)}} \\
\hline
\textbf{Yield(p)} & 
$\displaystyle (U_{11},V_{11})$ &  
$\displaystyle (U_{12},V_{12})$ \\
\hline
\textbf{Merge(1-p)} & 
$\displaystyle (U_{21},V_{21})$ & 
$\displaystyle (U_{22},V_{22})$ \\
\hline
\end{tabularx}
\end{table}
where $ p $ represents the probability that AV chooses to yield and $ q$ represents the probability that MV chooses to yield. The pair $ (U,V)$ indicates the payoff for AV and MV. 
\subsection{Online Driving style Estimation}
The driving style determines the weight of each component in the payoff function. Vehicles with different driving styles may adopt different strategies when facing the same scenario. Therefore, it is necessary to actively interact in order to estimate the driving style of MVs in real-time, enabling a more accurate evaluation of their payoff functions. 

In game theory, the equilibrium ultimately reached represents the optimal responses perceived by both players. This allows predicting strategies of opponent game player through equilibrium solutions. For merging scenarios, the payoff function of MV can be parameterized by its driving behaviors and states where the driving style weight $ \omega(k)$ critically influences ESS computation.
However, since $ \omega(k)$ cannot be directly observed by the AV, it necessitates coupling predicted behaviors with observed behaviors for online dynamic evaluation. Key assumptions for merging include:

\textit{Assumption 1}: Before entering the merging zone, the AV has no prior knowledge of human drivers' driving style $ \omega(k)$, but it is known that $ \omega(k) \in (0,1)$.

\textit{Assumption 2}: The human driver's driving style is deterministic and invariant to environmental changes.

This work proposes a real-time driving style estimation algorithm based on Bounded Rationality Modeling and ESS, as shown in Algorithm \ref{algorithm 3}. The algorithm employs replicator dynamics to model the bounded rationality of MVs. By solving ESS and its stability condition $ \omega(k_j)\in(ess\_lowerbounds , ess\_upperbounds)$, it predicts the optimal response of the MV and compares with observed behaviors. This process iteratively updates the upper and lower bounds of $\omega(k_j) $ until convergence is achieved.

\begin{algorithm}[t]
\caption{Driving Style Estimation}
\label{algorithm 3}
\begin{algorithmic}[1]
\Require Movement data under Evolutionary Stable Strategy $(p^*,q^*)$: $v_j$
\Ensure the driving style estimation of $MV_j$: $\omega(k_j)$
\State Initialize the lower and upper bounds: $k_l = 0$, $k_u = 1$ and $\omega(k_j) = (k_l + k_u) / 2$
\If {$q^* = 0$ and $v_j(t) \leq v_j(t-1)$}
    \State $k_u = \text{ess\_lowerbounds}(p^*, q^*)$
\ElsIf {$q^* = 1$ and $v_j(t) > v_j(t-1)$}
    \State $k_l = \text{ess\_upperbounds}(p^*, q^*)$
\EndIf
\State Update $\omega(k_j) = (k_l + k_u) / 2$
\end{algorithmic}
\end{algorithm}

In this process, the decision corresponding to $ q^*$ is treated as the estimated reaction. The observed behaviors derived from monitoring the velocity changes of MV between time steps $ t-1$ and $ t$. Here, if the AV predicts that MV should decelerate $ (q^*=0)$, but the observed velocity of MV does not increase $ v_j(t) \leq v_j(t-1)$, it indicates the current estimation is too conservative, thus updating the upper bound of $ \omega(k)$ to shift the driving style estimate toward a more aggressive direction, and vice versa.The algorithm iteratively updates bounds until convergence to a sufficiently small interval. Notably, the observed maneuver is related to the previous time points $ t-1$ (i.e. line 2 to line 5 in Algorithm \ref{algorithm 3}). Therefore, the algorithm must be performed chronologically. After driving style is estimated, we can update payoff of MVs and solve the ESS more accurately as described in Section C and D.

\subsection{Multi-objective Payoff Function}
To construct the game matrix, we introduced the payoff function $ J_i$ of vehicle $ i$ to calculate $ (U,V)$, which is a linear combination of efficiency, comfort and safety.
\begin{equation}
\begin{aligned} J_i = \omega(k)  J_\textit{efficiency,i} + (1 - \omega(k))  J_\textit{comfort,i} + \omega_s  J_\textit{safety,i} \end{aligned},
\end{equation}
where $ \omega(k)$ represents the driver's driving style, serving as a weight for different payoff terms. The more aggressive a driver's driving style, the greater his emphasis on driving efficiency. $ \omega_s$ denotes the conflict penalty coefficient. It is designed to prevent AV and MV from making identical decisions simultaneously.

Next, we solve the payoff of efficiency $ J_\textit{efficiency,i}$.
\begin{equation}
\begin{aligned} J_\textit{efficiency,i}= t_i(v_{-i}, d_{-i}, T) = \begin{cases}  \frac{d_{-i}}{v_{-i}} + T& \quad p=0 \\ \frac{d_{-i}}{v_{-i}} - T & \quad p=1 \end{cases} \end{aligned},
\end{equation}
where $ t_i $ denotes the travel time from the current position to the intersection point of road centerline. $ v_{-i} $ represents the speed of the opponent game player. $ d_{-i} $ is the distance of opponent game player to the intersection point of road centerline. $ T $ indicates the safe \textit{Time Headway} on the main road. When one player chooses to cut-in, he needs to arrive at the intersection point of road centerline $ T$ seconds earlier than the other. If he chooses to yield, the right-of-way priority is reversed.

Then we define the comfort payoff $ J_\textit{comfort,i}$.
\begin{equation}
\begin{aligned} J_\textit{comfort,i}   = \bar{a}_i^2(v_i,d_i,t_i) = \left[ \frac{2(d_i - v_i t_i)}{t_i^2} \right]^2 \end{aligned},
\end{equation}
where $ \bar{a}_i$ is the average acceleration required to reach the intersection point. $ v_i$ and $ d_i$ refer to the speed and distance to the intersection point of road centerline.

Last we solve the payoff of safety $ J_\textit{safety,i}$. We calculate the safety cost by examining vehicle collisions.
\begin{equation}
 J_\textit{safety,i}= J_{s}\left(\bar{a}_{i}, \bar{a}_{-i}\right) =|\bar{a}_i + \bar{a}_{-i}|,
\end{equation}
where $ \bar{a}_{-i}$ is the average acceleration of the opponent game player. 

In the merging game-theoretic model, the payoff function integrates conflict penalty weights to both simultaneous competitive acceleration and excessive passive deceleration. When two players have conflicting operational decisions (e.g., both choosing to accelerate for merging), the conflict penalty weight $ \omega_s$ set to 1, indicating a collision risk penalty is applied; otherwise, $ \omega_s = 0$.

\subsection{The Evolutionary Game Solver}
Now, we need to solve the proposed game which is described in the previous section. Unlike classical matrix games, EGT replaces NE with ESS. The following section explains how to solve for ESS using \textit{replicator dynamic equation}. Drawing on system dynamics principles, an evolutionary game can be viewed as a dynamical system where the strategy pair $(p,q) $ serves as internal parameters, and their evolution is governed by \textit{replicator dynamic equation}:
\begin{subequations}
\begin{flalign}
&& E_\textit{AV} &= p\mathbb{E}_\textit{AV1} + (1-p)\mathbb{E}_\textit{AV2}, & \\  
  && E_\textit{MV} &= q\mathbb{E}_\textit{MV1} + (1-q)\mathbb{E}_\textit{MV2}, & \\  
  && p(t) &= \frac{dp}{dt} = p(\mathbb{E}_\mathrm{AV1} - \mathbb{E}_\mathrm{AV}), & \\
  && q(t) &= \frac{dq}{dt} = q(\mathbb{E}_\mathrm{MV1} - \mathbb{E}_\mathrm{MV}), 
\end{flalign}
\end{subequations}
where the payoffs can be calculated:
\begin{subequations}
\begin{flalign}
&&\mathbb{E}_\textit{AV1}&=qU_\textit{11}+(1-q)U_\textit{12},& \\
&&\mathbb{E}_\textit{AV2}&=qU_\textit{21}+(1-q)U_\textit{22},& \\
&&\mathbb{E}_\textit{MV1}&=pV_\textit{11}+(1-p)V_\textit{21},& \\
&&\mathbb{E}_\textit{MV2}&=pV_\textit{12}+(1-p)V_\textit{22}.&
\end{flalign}
\end{subequations}

Based on the above \textit{replicator dynamic equation}, the ESS of the game can be calculated. The necessary conditions for ESS are as follows:
\begin{subequations}
\begin{flalign}
F(p) = \frac{dp}{dt} &= p(1-p)(E_\mathrm{AV1} - {E_\mathrm{AV2}}) = 0 \label{F(p)=0}, \\
F(q) = \frac{dq}{dt} &= q(1-q)(E_\mathrm{MV1} - {E_\mathrm{MV2}}) = 0 \label{F(q)=0},\\
\frac{\partial F(p)}{\partial p} &< 0 ,\frac{\partial F(q)}{\partial q} < 0.
\end{flalign}
\end{subequations}

According to Eq. (\ref{F(p)=0}) and Eq. (\ref{F(q)=0}), the pure strategy equilibrium points of the system can be calculated as: $ E_1(0,0), E_2(0,1), E_3(1,0),E_4(1,1)$. In EGT, an ESS may exist as a pure strategy equilibrium or be absent entirely. Consequently, it is necessary to evaluate the stability of pure strategy equilibrium points. This is achieved using Lyapunov first method by solving the system's Jacobian matrix.

\begin{equation}
\renewcommand{\arraystretch}{1.5}
\begin{aligned} 
J|_{(p,q)=E_i} = \left[ \begin{array}{cc} 
\frac{\partial F(p)}{\partial p} & \frac{\partial F(p)}{\partial q} \\ 
\frac{\partial F(q)}{\partial p} & \frac{\partial F(q)}{\partial q} 
\end{array}\right]  = 
\left[ \begin{array}{cc} 
\lambda_1 & 0 \\ 
0 & \lambda_2 
\end{array}\right] 
\end{aligned}.
\end{equation}

If $\lambda_1,\lambda_2 <0$, the pure strategy equilibrium point corresponds to the $ E(p^*,q^*)$. If $ p^*<q^*$, the AV will decide to merge, predicting that the MV will decelerate and yield. Based on this decision, the control outputs are generated to execute the merging maneuver:
\begin{equation}
u(k)=\frac{2(d_i - v_i t_i)}{t_i^2},
\end{equation}
where $t_i$ can be calculated:
\begin{equation}
t_i=\frac{d_{-i}}{v_{-i}} - T.
\end{equation}

The second-order linear differential equation can be used to describe the longitudinal motion of vehicles, where acceleration is the control variable. Eq. (\ref{kinematics model}) shows its discrete form:
\begin{equation}
\begin{aligned}\begin{bmatrix} x(k+1) \\ v(k+1) \end{bmatrix}&= \begin{bmatrix} 1 & T_s \\ 0 & 1 \end{bmatrix}\begin{bmatrix} x(k) \\ v(k) \end{bmatrix}+ \begin{bmatrix} 0 \\ T_s \end{bmatrix}u(k).\end{aligned}
\label{kinematics model}
\end{equation}

By incorporating the vehicle kinematics model, the motion trajectory of AV is planned over the future time step $ k$.

\section{Numerical Analysis and Model Assessment}
The performance of the proposed game-theoretic decision making framework is evaluated in various scenarios and by different subjects. First, we consider the on-ramp merging scenario shown in Fig.\ref{scenario} and select the baseline methods. Then we design comprehensive tests using different driving styles of MVs to validate the effectiveness of the merging algorithm and the accuracy of the driving style estimation algorithm. Last, we compare our approach by different metrics.
\subsection{Implementation Details}
In the merging scenario, the AV is positioned on the ramp, driven by the EGT decision-making proposed in this paper, while five MVs with distinct driving styles are simulated on the main road using the IDM (reactive mode).The AV must decide which gap to merge before entering the convergence area, and then complete the lane change in the convergence area. Each test has a duration of $ T_{tr}$ = 10 s with a discretization step of $ \Delta t$ = 100 ms, and the number of timestamps is $ N_{ts}=T_{tr}/ \Delta t$. All simulations were conducted on a workstation equipped with an Intel Core i7-14700KF processor, 32 GB of RAM, and an NVIDIA GeForce RTX 3070 GPU.

\subsubsection{Game-Theoretic Behavior Planner} We use a planning horizon of $ T_{bp}$ = 25, a discretization step of $ t_{bp}$ = 0.2 s, a decision time period of $ h$ = 1 s, and a decision horizon of $ H$ = 5. In other words, the behavior planner looks ahead for 5 s and makes decision every 1 s. We run the behavior planner at 5 Hz.

We select the AV as the target vehicle. The vehicle behind the target vehicle plays a game with AV. Through game-theoretic interactions, the vehicles dynamically determine their relative positions and merging priorities, ultimately reaching a ESS that balances safety and efficiency.

\subsubsection{IDM models with different driving style parameters} We adjust the key parameters of IDM to simulate different driving styles, such as \textit{Time Headway} $ T$. Conservative drivers adopt a larger time headway to maintain a safe following distance, while aggressive drivers shorten the spacing between vehicles to improve traffic efficiency. 

\subsubsection{MPC lane change controller} We use MPC to control AV after passing the merge point \cite{kabzan2020amz}. The AV treats the vehicles ahead and behind as dynamic obstacles based on a pre-determined merging sequence, ultimately completing the lane change maneuver.

\subsection{Case Study}

\begin{table}[t]
\caption{SCENARIOS WITH DIFFERENT SETTINGS}
\centering
\begin{tabular*}{\linewidth}{@{\extracolsep{\fill}}lrcrrr@{}}
\toprule
\multicolumn{3}{c}{\textbf{Vehicle States}} & \multicolumn{3}{c}{\textbf{Time Headway (s)}} \\
\cmidrule(lr){1-3} \cmidrule(lr){4-6}
\textbf{ID} & \multicolumn{1}{r}{\textbf{$ d_i$ (m)}} & \textbf{$ v_i$ (m/s)} & \multicolumn{1}{c}{\textbf{I}} & \multicolumn{1}{c}{\textbf{II}} & \multicolumn{1}{c}{\textbf{III}}\\
\midrule
MV1 & 173.2 & 10 & \multicolumn{1}{c}{2} & \multicolumn{1}{c}{2} & \multicolumn{1}{c}{$\mathcal{N}(1,\,\sigma^2)$}\\
MV2 & 147.4 & 10 & \multicolumn{1}{c}{2} & \multicolumn{1}{c}{2} & \multicolumn{1}{c}{$\mathcal{N}(1,\,\sigma^2)$}\\
MV3 & 121.6 & 10 & \multicolumn{1}{c}{$\mathcal{N}(1,\,\sigma^2)$} & \multicolumn{1}{c}{$\mathcal{N}(2,\,\sigma^2)$} & \multicolumn{1}{c}{$\mathcal{N}(1,\,\sigma^2)$}\\
MV4 & 85.5 & 10 & \multicolumn{1}{c}{1} & \multicolumn{1}{c}{1} & \multicolumn{1}{c}{$\mathcal{N}(1,\,\sigma^2)$}\\
MV5 & 70.0 & 10 & \multicolumn{1}{c}{2} & \multicolumn{1}{c}{2} & \multicolumn{1}{c}{$\mathcal{N}(1,\,\sigma^2)$}\\
\bottomrule
\end{tabular*}
\label{tab2}
\end{table}

\graphicspath{{Merging decision-making based on Game-theory/includegraphics/pic_pdf_0427/}} 

\begin{figure*}[!htbp] 
  \centering
  \begin{subfigure}[t]{0.245\textwidth}
    \includegraphics[width=\linewidth, height=3cm]{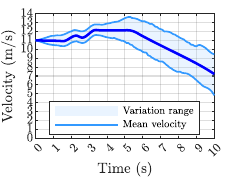}
    \caption{Velocity (S-I)}\label{v1}
  \end{subfigure}
\hfill
    \begin{subfigure}[t]{0.245\textwidth}
    \includegraphics[width=\linewidth, height=3cm]{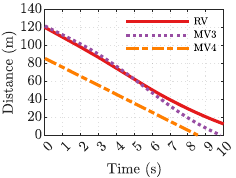}
    \caption{Distance (S-I)}\label{d1}
  \end{subfigure}
\hfill
\begin{subfigure}[t]{0.245\textwidth}
    \includegraphics[width=\linewidth, height=3cm]{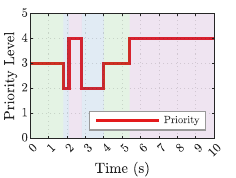}
    \caption{Sequence (S-I)}\label{s1}
  \end{subfigure}
\hfill
\begin{subfigure}[t]{0.245\textwidth}
    \includegraphics[width=\linewidth, height=3cm]{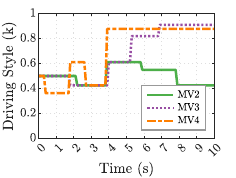}
    \caption{Driving style (S-I)}\label{k1}
  \end{subfigure}
\hfill
  
  \vspace{0.5em}
  
  \begin{subfigure}[t]{0.245\textwidth}
    \includegraphics[width=\linewidth, height=3cm]{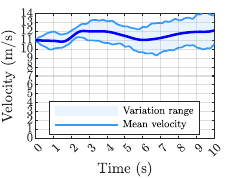}
    \caption{Velocity (S-I)}\label{v2}
  \end{subfigure}
\hfill
    \begin{subfigure}[t]{0.245\textwidth}
    \includegraphics[width=\linewidth, height=3cm]{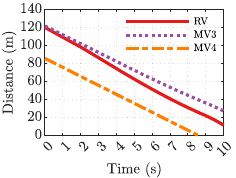}
    \caption{Distance (S-I)}\label{d2}
  \end{subfigure}
\hfill
\begin{subfigure}[t]{0.245\textwidth}
    \includegraphics[width=\linewidth, height=3cm]{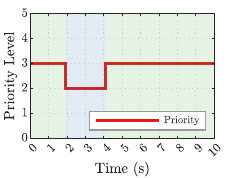}
    \caption{Sequence (S-I)}\label{s2}
  \end{subfigure}
\hfill
\begin{subfigure}[t]{0.245\textwidth}
    \includegraphics[width=\linewidth, height=3cm]{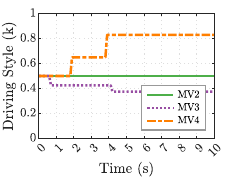}
    \caption{Driving style (S-I)}\label{k2}
  \end{subfigure}
\hfill
  \vspace{0.5em}
  
  \begin{subfigure}[t]{0.245\textwidth}
    \includegraphics[width=\linewidth, height=3cm]{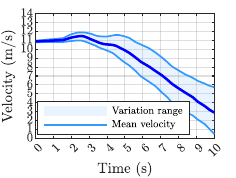}
    \caption{Velocity (S-I)}\label{v3}
  \end{subfigure}
\hfill
    \begin{subfigure}[t]{0.245\textwidth}
    \includegraphics[width=\linewidth, height=3cm]{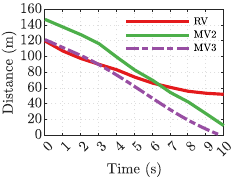}
    \caption{Distance (S-I)}\label{d3}
  \end{subfigure}
\hfill
\begin{subfigure}[t]{0.245\textwidth}
    \includegraphics[width=\linewidth, height=3cm]{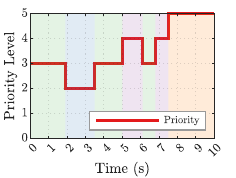}
    \caption{Sequence (S-I)}\label{s3}
  \end{subfigure}
\hfill
\begin{subfigure}[t]{0.245\textwidth}
    \includegraphics[width=\linewidth, height=3cm]{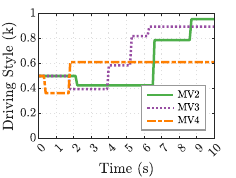}
    \caption{Driving style (S-I)}\label{k3}
  \end{subfigure}
\hfill

\caption{ Comparative simulation results between Scenario I: (a)-(d), Scenario II: (e)-(h) and Scenario III: (i)-(l) showing four key metrics: velocity profiles of AV, distance relationships between AV and MVs, sequence patterns of merging, and estimated driving styles of MVs respectively.}
\label{Simulation result}
\end{figure*}

To validate the effectiveness of the merging algorithm and the accuracy of the driving style estimation algorithm, we design three scenarios with identical vehicle parameters including distance to the merge point and speed but different \textit{Time Headway} settings, as shown in Table \ref{tab2}. Based on the normal distribution of \textit{Time Headway} in NGSIM data \cite{tang2021atac}, we designed \textit{Time Headway} in scenarios to follow two normal distributions: $ \mathcal{N}(1,\sigma^2)$ and $\mathcal{N}(2,\sigma^2)$ where $ \sigma = 0.5 $, representing aggressive and conservative driving styles respectively. The simulation results are illustrated in Fig. \ref{Simulation result}.

In the experiments, the AV successfully identifies different driving styles of main-road vehicles through active acceleration testing, as shown in \cref{v1,v2,v3},
and chooses to merge into the gap in front of more conservative vehicles, as illustrated in \cref{d1,s1,d2,s2,d3,s3}. This is evidenced by the distinct driving styles captured in \cref{k1,k2,k3}.

\begin{figure}[thpb]
  \begin{subfigure}[t]{0.24\textwidth}
    \includegraphics[width=\linewidth, height=3cm]{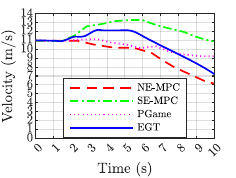}
    \caption{}\label{v_compare_AV}
  \end{subfigure}
\hfill
    \begin{subfigure}[t]{0.24\textwidth}
    \includegraphics[width=\linewidth, height=3cm]{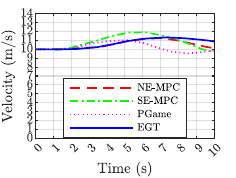}
    \caption{}\label{v_compare_MV3}
\end{subfigure}
\caption{ Comparative simulation results in Scenario I generated by EGT and other three baseline methods. (a): Velocity profiles of AV. (b): Velocity profiles of MV3 which has the highest level of interaction with the AV during overall tests.}
\end{figure}

In Scenario I, our experimental results demonstrate that during the merging process on ramp, the AV proactively accelerated at about \textit{$ t=2\ s$}. After a 0.5-second reaction delay, the trailing vehicle MV3 accelerated to reject intention of AV to merge. Simultaneously, AV observed real-time response of MV3, updated its driving style assessment through online estimation, and re-engaged in game-theoretic interactions to revise merging priority. Ultimately, AV completed the evaluation of adjacent vehicles' driving styles and opted to merge ahead of MV2, a more conservative driver positioned farther back, rather than merging ahead of the aggressive MV3.

In Scenario II, where MV3 is predefined as a conservative driver, AV dynamically accelerated to merge ahead of it despite similar initial positions. In Scenario III, where all MVs are predefined as aggressive drivers, AV kept slowing down until it finally merged into the main road. This highlights the proactive adaptability of our proposed decision-making algorithm. Notably, such strategies align more closely with human drivers' behavioral expectations, thereby improving acceptance.

\subsection{Statistical Results}
We compare the proposed method with three baseline methods across 100 tests from scenario I.
\begin{itemize}
\item NE-MPC\cite{zhang2023efficient}, combining a NE-seeking behavior planner with a MPC motion planner. This method does not consider merging scenario as a dynamic game.
\item SE-MPC\cite{zhang2019game}, which searches for the Stackelberg equilibrium as final strategy and uses MPC for motion planning, which formulates interactions between the AV and MVs as a Stackelberg game.
\item PGame\cite{liu2023potential}, a game-theoretic method using sampling-based trajectories as actions.
\end{itemize}

The quality of the trajectories is evaluated using the following metrics.
\begin{itemize}
\item Jerk, since we assess the driver's acceptance via human drivers' comfort\cite{irshayyid2023comparative}. The human drivers' comfort is affected by sudden changes in acceleration, known as jerk (the rate of change of acceleration).
\item Terminal speed, which significantly impacts drivers' acceptance of autonomous systems. Merging maneuver by the AV can delay the traffic flow speed downstream. Therefore, the terminal speed of MV5 serves as a key metric to evaluate the impact of merging maneuver on main-road traffic efficiency.
\item Collision rate and time-to-collision (TTC), which can evaluate the safety level, is also important metric to drivers' acceptance. The collision rate is defined as the proportion of the number of cases. And TTC refers to the time MV takes to collide if they maintain their current speed and heading.
\end{itemize}

Previous studies have indicated that these factors significantly influence human drivers' trust and acceptance of autonomous vehicles \cite{tang2021atac,yuan2016suppose}. Therefore, improvements in these metrics enhance the interaction comfort and social acceptance between AVs and human-driven vehicles. Accordingly, we recorded the positions, velocities, and accelerations of all vehicles during the merging process and calculated the following metrics: mean jerk, maximum jerk, terminal speed of MV5, collision rate, and mean TTC. The results are summarized in Table \ref{tab:performance}.

\begin{table}[t]
\centering
\caption{PERFORMANCE COMPARISON}
\begin{tabular}{lcccc}
\toprule
\textbf{Metric}  & \textbf{NE-MPC} & \textbf{SE-MPC} & \textbf{PGame} & \textbf{EGT}\\
\midrule
Mean jerk ($\mathrm{m/s^3}$)  & 0.31 & 0.31 & 0.42 & \textbf{0.25} \\ 
Max jerk ($\mathrm{m/s^3}$)  & 0.58 & 0.61 & 4.77 & \textbf{0.52} \\
Terminal speed ($\mathrm{m/s}$)  & 8.75 & 8.41 & 7.88 & \textbf{9.12} \\
Collision rate (\%)  & \textbf{0} & 1 & 6 & \textbf{0} \\
Mean TTC (s)  & 4.51 & 5.85 & 5.45 & \textbf{6.22} \\
\bottomrule
\end{tabular}
\label{tab:performance}
\end{table}

The analysis reveals that we reduces 18.3$ \%$ of mean jerk and 10.3$ \%$ maximum jerk of MVs, lowers the collision rate, and improves 4$ \%$ of the terminal speed of MV5. However, as hypothesized (Assumption 1), the game-theoretic interactions initiated in advance perturb the main-road platoon earlier, leading to an 6.3$ \%$ increase in mean TTC. A higher TTC means MV drives safer but may decrease the efficiency.

In summary, our proposed merging decision-making algorithm improves traffic flow efficiency while enhancing ride comfort and safety. By effectively mitigating the disruptive impact of merging maneuvers on MVs, the algorithm demonstrates potential for improved social acceptance.
\section{Conclusions}

In this study, an EGT framework for autonomous merging decision-making is developed, integrating real-time driving style estimation. By formulating interactions as bounded rationality games and dynamically adjusting payoff functions through observed MV reactions, the framework achieves human-like decision-making while balancing benefits for all vehicles and enhancing social acceptance. Experimental results demonstrate that the proposed method outperforms existing game-theoretic and traditional planning approaches, achieving smoother maneuvers, higher traffic efficiency, and improved adaptability to heterogeneous driving styles. The online driving style estimation mechanism further refines decision-making by continuously aligning AV behaviors with human expectations. These advancements demonstrate the framework’s potential to align decision-making algorithms with human driving behaviors, paving the way for more socially aware and acceptable autonomous driving systems. Future research will extend this framework to cooperative multi-lane scenarios through Vehicle-to-Infrastructure (V2I) coordination, further bridging the gap between game-theoretic models and practical autonomous driving systems.

\section*{Acknowledgment}
This work was supported by the National Natural Science Foundation of China under Grant 52472444. (Co-Corresponding author: Huilong Yu and Junqiang Xi.)

\bibliographystyle{IEEEtran}  
\bibliography{IEEEabrv, ref}  


\end{document}